\documentclass{article}

\usepackage{microtype}
\usepackage{graphicx}
\usepackage{subcaption}
\usepackage{booktabs} 
\usepackage{multirow} 

\usepackage{hyperref}

\usepackage{amd}

\usepackage{amsmath}
\usepackage{amssymb}
\usepackage{mathtools}
\usepackage{amsthm}

\usepackage{graphicx}
\usepackage{booktabs}
\usepackage[table]{xcolor}
\usepackage{tcolorbox}
\usepackage{tabularx} 
\usepackage{array}
\usepackage[labelfont=bf,justification=justified]{caption}

\definecolor{mycolor}{RGB}{230,220,250}
\definecolor{mycolor2}{RGB}{240,235,252}
\definecolor{mygold}{RGB}{193,169,104}

\usepackage[capitalize,noabbrev]{cleveref}
\usepackage{tikz}
\usepackage{xspace}
\usepackage{xspace}
\usepackage{wrapfig}

\newcommand{\AMDlogo}{\includegraphics[height=0.28in]{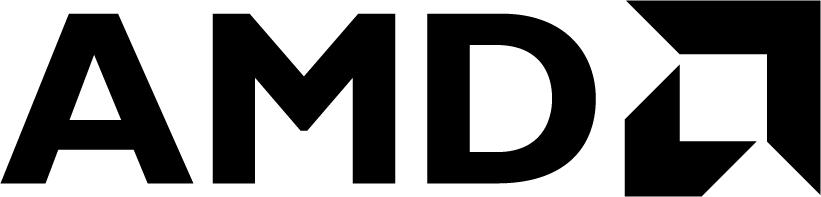}}

\theoremstyle{plain}

\theoremstyle{definition}

\theoremstyle{remark}

\usepackage[textsize=tiny]{todonotes}

\amdtitlerunning{DC-DiT: Adaptive Compute and Elastic Inference for Visual Generation via Dynamic Chunking}

\begin{document}

\onecolumn

  \amdtitle{DC-DiT: Adaptive Compute and Elastic Inference for Visual Generation via Dynamic Chunking}



  \icmlsetsymbol{equal}{*}

  \begin{amdauthorlist}
    \amdauthor{Akash Haridas
    }{equal,amd,cor}
    \amdauthor{Utkarsh Saxena}{amd}
    \amdauthor{Parsa Ashrafi Fashi}{amd} \\
    \amdauthor{Mehdi Rezagholizadeh}{amd,cor}
    \amdauthor{Vikram Appia}{amd}
    \amdauthor{Emad Barsoum}{amd}
  \end{amdauthorlist}
  
  \amdaffiliation{amd}{Advanced Micro Devices Inc. (AMD)}

  \amdcorrespondingauthor{Akash Haridas}{akash.haridas@amd.com}
  \amdcorrespondingauthor{Mehdi Rezagholizadeh}{mehdi.rezagholizadeh@amd.com}

  \amdkeywords{Machine-Learning, Image-Generation}

  \vskip 0.3in



\amdPrintAffiliationsInline

\begin{abstract}

Diffusion Transformers rely on static \textit{patchify} tokenization, assigning the same token budget to smooth backgrounds, detailed object regions, noisy early timesteps, and late-stage refinements. We introduce the \emph{Dynamic Chunking Diffusion Transformer} (DC-DiT), which replaces fixed patchification with a learned encoder-router-decoder scaffold that adaptively compresses the 2D input into a shorter token sequence through a chunking mechanism learned end-to-end with diffusion training. DC-DiT allocates fewer tokens to predictable regions and noisy timesteps, and more tokens to detailed regions and later refinement stages, yielding meaningful spatial segmentations and timestep-adaptive compression schedules without supervision. Furthermore, the router provides an importance ordering over retained tokens, enabling \emph{elastic inference}: a single checkpoint can be evaluated at flexible compute budgets with a smooth quality-compute tradeoff. Additionally, DC-DiT can be \emph{upcycled} from pretrained DiT checkpoints and is also compatible with orthogonal dynamic computation approaches. On class-conditional ImageNet generation, DC-DiT reduces inference FLOPs by up to $36.8\%$ and improves FID by up to $37.8\%$ over DiT baselines, yielding a stronger quality--compute Pareto frontier across model scales, resolutions, and guidance settings. More broadly, these results suggest that adaptive tokenization is a general mechanism for making visual generation both more efficient and more flexible at inference time.

\end{abstract}

\section{Introduction}

Transformer-based diffusion models \cite{peebles2023scalablediffusionmodelstransformers} achieve strong image generation quality, but they commonly rely on fixed tokenization schemes: a fixed \textit{patchify} operation converts every image into the same grid of tokens at every denoising step. This design makes token count an architectural constant. A uniform background patch, a high-frequency object boundary, an early noisy timestep, and a late detail-refinement timestep all receive the same token budget. This ignores two sources of natural adaptivity in image generation: different spatial regions contain different amounts of detail, and diffusion trajectories typically progress from coarse to fine structure across timesteps.

Recent work has explored dynamic computation as a way to reduce redundancy in diffusion transformers. These methods typically adapt computation after a fixed token sequence has already been formed, for example by pruning or dropping less informative tokens \cite{wang2024dynamicdiffusiontransformer}, merging redundant tokens \cite{wu2025importancebasedtokenmergingefficient}, reducing intermediate feature dimensions \cite{wang2024dynamicdiffusiontransformer}, or reusing hidden states across nearby denoising steps \cite{liu2025timestepembeddingteacache}. Such approaches have shown that substantial computation can be saved along both spatial and temporal axes with limited degradation in generation quality. However, they generally operate within a fixed-tokenization regime: the initial \textit{patchify} operation still maps every image into the same regular grid of tokens, independent of image content or denoising timestep. Thus, while these methods reduce computation inside or around the transformer backbone, they do not address the more fundamental limitation that the token sequence itself is statically defined before the model begins processing.

\begin{figure}[t]
\centering
\begin{minipage}[c]{0.50\textwidth}
    \centering
    \includegraphics[width=\linewidth]{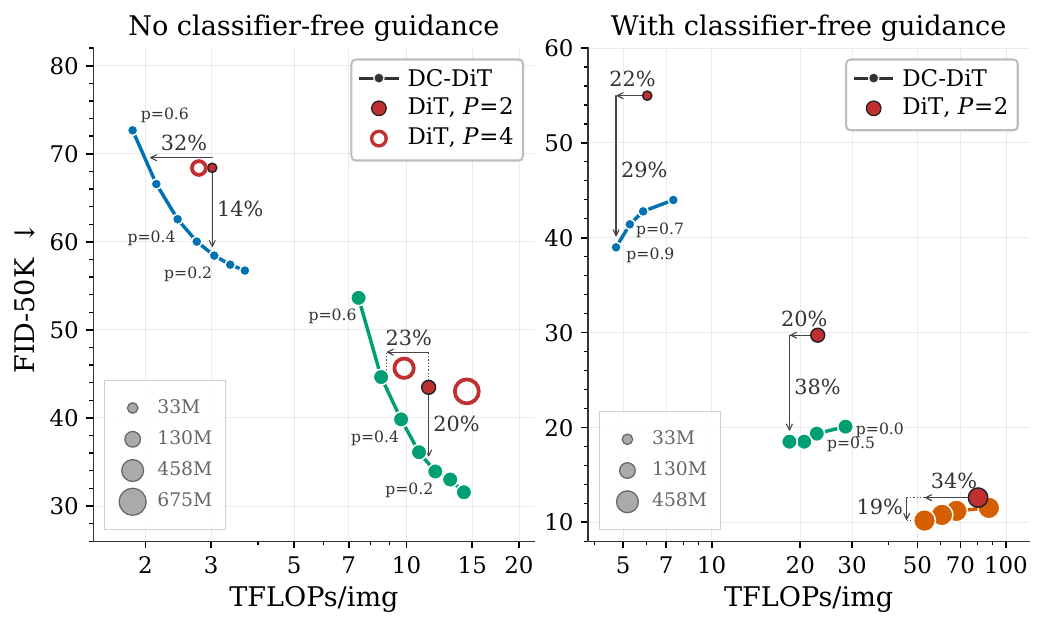}
\end{minipage}
\hfill
\vrule width 0.4pt
\hfill
\begin{minipage}[c]{0.46\textwidth}
    \centering
    \includegraphics[width=\linewidth]{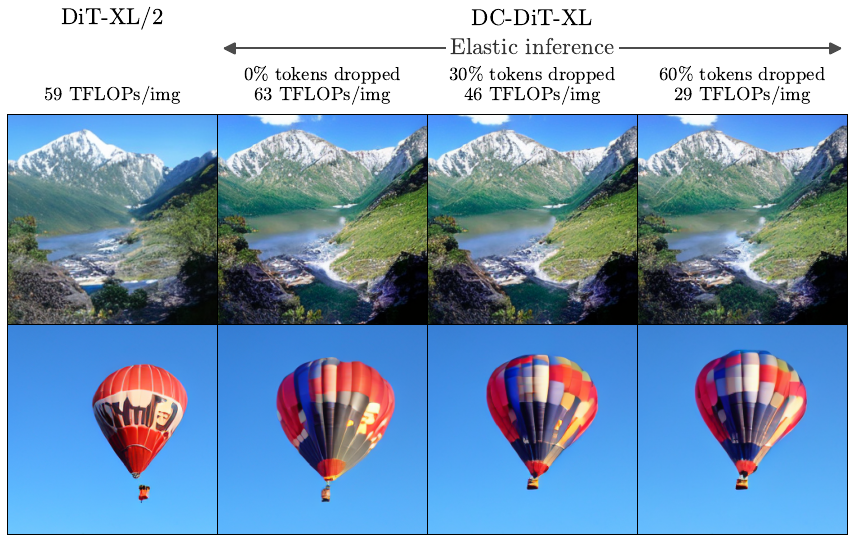}
\end{minipage}
\caption{Left: By learning to re-allocate compute across spatial regions and timesteps, DC-DiT improves the FID--FLOPs Pareto frontier of DiT. Right: DC-DiT enables elastic inference, where a single trained checkpoint can be evaluated at multiple inference budgets while preserving visual quality.}
\label{fig:main_results}
\end{figure}

To address this limitation, we introduce the Dynamic Chunking Diffusion Transformer (DC-DiT), which replaces fixed patchification with an encoder--router--decoder scaffold learned end-to-end through diffusion training. Rather than processing every latent position uniformly, DC-DiT selects a compact set of informative tokens for transformer computation and reconstructs the dense diffusion prediction afterward. This makes tokenization adaptive: for each input and denoising timestep, the model learns where it should allocate its token budget.

This adaptive routing mechanism learns meaningful spatial and temporal allocation patterns without explicit supervision. Spatially, smooth background regions and low-variation areas are compressed into fewer tokens, while object regions, boundaries, and high-frequency details are represented with denser token allocations. Temporally, the model naturally uses fewer tokens at noisier denoising steps and progressively increases the token budget as the sample becomes cleaner, mirroring the coarse-to-fine structure of the diffusion trajectory. We further introduce a multi-budget training objective that enables \textit{elastic inference}: a single trained checkpoint can be evaluated at multiple token budgets, allowing a smooth quality--compute curve at test time. This also enables \emph{Lite-CFG}, which spends more compute on the conditional branch and more aggressively compresses the unconditional branch during classifier-free guidance (CFG).

Beyond training from scratch, DC-DiT provides a practical path for upgrading existing DiT models: a pretrained fixed-patch DiT can be upcycled into DC-DiT with lightweight finetuning, retaining the benefits of pretraining while enabling adaptive tokenization. Since DC-DiT addresses token allocation before backbone computation, it remains orthogonal to post-hoc efficiency techniques such as token merging, token pruning, hidden-dimension reduction, and timestep caching and can be combined with them to enhance benefits. As a result, DC-DiT meaningfully improves the FID--FLOPs Pareto frontier across model scales, guidance settings, and resolutions, and exposes a smooth quality--compute tradeoff from a single checkpoint. Our contributions can be summarized as follows:
\begin{itemize}
	\item We propose DC-DiT, a Diffusion Transformer that learns to adaptively compress the 2D input into a token sequence in a data-dependent manner with a mechanism learned end-to-end during diffusion training.
	\item Through multi-budget training, we enable elastic inference, which allows a single trained DC-DiT checkpoint to be evaluated at multiple token budgets, further enabling \emph{Lite-CFG}, which allocates asymmetric compute in classifier-free guidance.
  \item We evaluate DC-DiT across ImageNet model scales, resolutions, and guidance settings, showing meaningful improvements in the quality--compute Pareto frontier, with up to $36.8\%$ lower inference FLOPs and up to $37.8\%$ better FID than fixed-patch DiT baselines.
  \item We demonstrate that DC-DiT extends beyond class-conditional generation by upcycling Z-Image with lightweight adaptation, and show that DC-DiT composes well with orthogonal acceleration methods such as DyDiT and TeaCache.
\end{itemize}

\section{Related Work}
\label{sec:related_work}

\textbf{Compute-adaptive Diffusion Transformers.} Several methods introduce adaptivity directly into the DiT backbone architecture or training procedure. DyDiT~\cite{wang2024dynamicdiffusiontransformer} adapts hidden width across timesteps and prunes spatial tokens that are predicted to be less informative. D$^2$iT~\cite{jia2025d2itdynamicdiffusiontransformer} moves adaptivity into the latent representation, using a Dynamic VAE to encode different regions at different downsampling rates. Other approaches operate at the token-routing level. DiffCR~\cite{lin2024layertimestepadaptivedifferentiable} learns layer- and timestep-dependent compression ratios, SparseDiT~\cite{yang2024flexditdynamictoken} varies token density across network depth, and DiffMoE~\cite{zhou2025diffmoedynamictoken} combines token-level routing with mixture-of-experts capacity allocation. Alternative to learned adaptive architectures, many inference-time methods accelerate pretrained DiTs without additional training. Token merging~\cite{wu2025importancebasedtokenmergingefficient,bolya2022tome,bolya2023tomesd,fang2025sdtm} reduces sequence length by combining redundant tokens, while early-exit strategies~\cite{moon2024simpleearlyexit} skip unnecessary computation in later layers. Caching methods exploit the temporal redundancy of the denoising trajectory: TeaCache~\cite{liu2025timestepembeddingteacache}, BlockCache~\cite{wimbauer2024blockcache}, TaylorSeer~\cite{liu2025reusingtaylorseer}, ToCA~\cite{zou2024acceleratingtoca}, $\Delta$-DiT~\cite{chen2024deltadit}, and HarmoniCa~\cite{huang2024harmonica} reuse outputs, hidden states, token features, or feature differences across adjacent timesteps. Sparse-attention methods~\cite{yuille2025groupingfirstattending} reduce the cost of self-attention by limiting each token to a subset of keys and values, and structured-sparsity approaches such as Chipmunk~\cite{silveria2025chipmunk} and Just-in-Time spatial acceleration~\cite{sun2026justintime} further exploit activation- or token-level sparsity. Together, these works show that the computational budget of DiTs can be dynamically concentrated on the most useful regions, tokens, layers, and timesteps instead of being spent uniformly throughout the generation process.

\textbf{Content-Adaptive Tokenization.}
Beyond diffusion model architectures, a broad line of work studies tokenizers that adapt the number, size, or granularity of tokens to the input content. In vision and vision-language models, DynamicViT~\cite{zhou2021dynamicvitefficientvision} progressively prunes tokens using learned importance scores, while APT~\cite{choudhury2025acceleratingvisiontransformers} allocates variable-size patches according to local information content. Variable-length image encoders such as ALIT~\cite{duggal2024adaptivelengthimagetokenization}, ElasticTok~\cite{Yan2024ElasticTokAT}, and DOVE~\cite{mao2025imagesworthvariablelength} similarly emit content-dependent token counts, producing compact representations for simple regions and denser representations for more informative ones. Content-adaptive tokenization has also gained attention in language modeling. BLT~\cite{zhou2024bytelatenttransformer} groups raw bytes into variable-length patches using entropy-based heuristics, while SuperBPE~\cite{liu2025superbpespacetravel} learns cross-whitespace merges to reduce sequence length. More recent learned approaches such as H-Net~\cite{hnet} and DLCM~\cite{qu2026dynamiclargeconceptmodels} predict dynamic boundaries and route computation through compressed chunk- or concept-level representations. Together, these methods reduce redundancy by shifting computation from individual tokens toward higher-level, content-dependent units.

\section{Method}
\label{sec:method}
DC-DiT replaces the static patch grid with a shorter data-dependent sequence, making tokenization learnable through an encoder-router-decoder scaffold. In this section, we describe the architecture of DC-DiT and its training procedure.

\subsection{Overall architecture}
The standard DiT patchifies the input latent image into non-overlapping $P \times P$ patches, with $P>1$ fixed during both training and inference~\cite{peebles2023scalablediffusionmodelstransformers}. In contrast, DC-DiT operates on the flattened latent grid, equivalent to patching with $P=1$, and learns to dynamically group nearby latent pixels into content-dependent vision tokens jointly with diffusion training.

\begin{figure}[t]
\centering
\includegraphics[width=\textwidth]{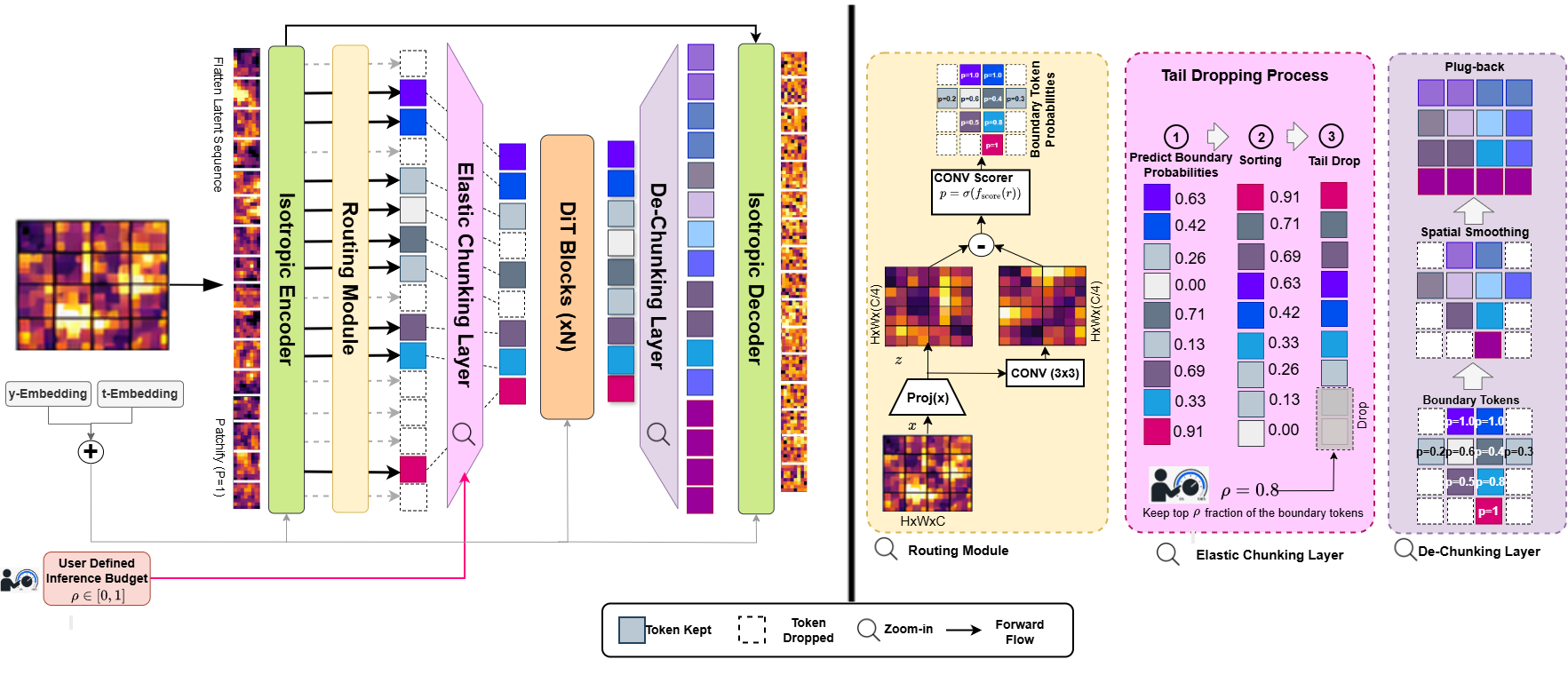}
\caption{ Architecture of DC-DiT. The isotropic encoder aggregates local context across the input tokens. The chunking layer selects a subset of boundary tokens via a learned routing module, yielding a compressed sequence that is processed by the DiT blocks. The de-chunking layer restores the original resolution through spatial smoothing followed by plug-back.}
\label{fig:Dc-DiT}
\end{figure}

DC-DiT wraps a DiT backbone with an encoder-router-decoder scaffold, illustrated in Figure~\ref{fig:Dc-DiT}. First, an isotropic encoder mixes local information across neighboring latent tokens, producing features suitable for routing. A router then predicts boundary probabilities: tokens that are hard to predict from their neighborhood are assigned high probabilities, while locally redundant tokens are assigned lower probabilities. The elastic chunking layer retains a shortened sequence of tokens based on boundary probabilities. The shortened tokens are further processed by a DiT backbone, which comprises a series of blocks with architecture similar to \cite{peebles2023scalablediffusionmodelstransformers}. Before the backbone, sinusoidal positional embeddings are added using each retained token's original 2D grid position. After the backbone, a de-chunking layer restores the sequence to the original latent-grid resolution, and an isotropic decoder maps it back to the diffusion prediction space. A residual connection from the encoder output is added after de-chunking and before decoding to preserve fine-grained spatial information. This residual is gated by the router's boundary probabilities using a straight-through estimator (STE)~\cite{bengio2013estimating}, which keeps discrete routing decisions in the forward pass while allowing gradients to flow through the routing probabilities during training. This inference procedure is repeated for each diffusion timestep.

Overall, DC-DiT consists of: (1) an encoder that prepares local features for routing, (2) a router that assigns boundary probabilities to tokens, (3) a chunking layer that keeps boundary tokens and drops predictable non-boundaries, (4) DiT blocks operating on the shortened sequence with original-position embeddings, (5) a de-chunking layer that reconstructs full resolution, and (6) a decoder that produces the diffusion-model prediction.

\subsection{Encoder and Decoder}
\label{sec:encoder-decoder}

\noindent
The encoder and decoder are isotropic modules that preserve the token count while mixing information over the 2D spatial grid. The encoder aggregates local context to support the router's boundary decisions, while the decoder maps the de-chunked features back to the diffusion model's prediction space. We instantiate these modules with convolutional residual blocks following~\cite{Rombach_2022_CVPR}. Each block reshapes the token sequence into a 2D feature map of shape $(H, W, D)$, applies two $3{\times}3$ convolutions with GroupNorm and SiLU activations, injects the conditioning vector after the first convolution, and adds a residual connection after the second. The result is then flattened back into a token sequence. Because the encoder and decoder operate before sequence compression and after decompression, respectively, their cost is non-negligible. To reduce overhead, both modules use an intermediate hidden width equal to one quarter of the main transformer dimension, projecting to the full dimension only at the encoder output for routing and projecting back down at the decoder input.

\subsection{Router and Elastic Chunking Layer}
\label{sec:chunking}

The chunking layer maps a full token sequence $X \in \mathbb{R}^{B \times L \times d}$ to a shorter sequence $X' \in \mathbb{R}^{B \times M \times d}$ by selecting a subset of tokens as \emph{boundary tokens}. These tokens define the retained representatives that are processed by the DiT backbone.

\textbf{Routing score and boundary probability.}
The router predicts a boundary probability $p_i \in [0,1]$ for each encoded token. Its design is based on local spatial predictability: after the encoder has mixed nearby information, tokens that are difficult to predict from their neighborhood should be retained, while locally predictable tokens can be dropped and later reconstructed from nearby representatives. Concretely, we reshape the encoded sequence into an $H \times W$ feature grid and project it to a bottleneck representation $\mathbf{z}$. A lightweight $3{\times}3$ convolutional predictor estimates each feature from its local spatial context, yielding $\hat{\mathbf{z}}$. The residual \(\mathbf{r} = \mathbf{z} - \hat{\mathbf{z}}\)
measures local unpredictability. A small convolutional scorer $f(\cdot)$ maps this residual to a logit, and the boundary probability at position $i$ is
\begin{equation}
p_i = \sigma\!\left(f_{\mathrm{score}}(\mathbf{r})_i\right).
\end{equation}

\textbf{Elastic Chunking Layer.} The chunking layer generates a hard boundary mask by thresholding $p_i > 0.5$. To maintain differentiability and enable end to end learning, we train with a straight through estimator (STE)~\cite{bengio2013estimating}. During the forward pass, chunking layer uses the hard boundary mask while during the backward pass, gradients are propagated as if the mask were the continuous probability $p_i$. Without any external supervision, we observe that high-probability boundary tokens concentrate around edges, textures, and salient regions, whereas low-probability tokens typically occur in locally smooth or predictable areas.

\textbf{Elastic inference via tail dropping.}
The router's boundary probabilities provide more than a binary keep/drop decision: they also induce a ranking of retained tokens from most to least important. We exploit this ranking to enable elastic inference through \emph{tail dropping}. Let \(B = \{i : p_i > 0.5\}\) be the natural boundary set selected by the router. For a user-specified tail-dropping fraction \(\rho \in [0,1)\), we sort the selected boundaries by \(p_i\) and drop the lowest \(\rho |B|\) of them before packing the sequence for the inner DiT blocks. The retained set \(B_\rho \subseteq B\) preserves the router's most confident representatives while increasing the effective compression ratio from \(L/|B|\) to \(L/|B_\rho|\). As a result, a single trained checkpoint can be evaluated at multiple compute budgets by changing only \(\rho\), without modifying model weights.

\textbf{Lite CFG.} During Classifier Free Guidance (CFG), each sampling step evaluates both conditional and unconditional branches. We introduce \emph{Lite-CFG}, which leverages DC-DiT's elastic budget control to assign a conservative tail-dropping fraction to the conditional branch and a larger drop fraction to the unconditional branch. This spends most of the token budget on the branch that carries class information, while reducing the cost of the unconditional prediction used for guidance.

\textbf{Batched chunking and sequence packing.}
During batched training and inference, each sample may have a different number of boundary tokens \(M\) selected by the router. To enable efficient processing and avoid wasting compute from padding, we apply sequence packing: the valid boundary tokens from all samples are concatenated along the sequence dimension and the per-sample boundaries are handled by the variable-length attention kernel of FlashAttention~\cite{dao2022flashattention}. The pointwise components of the inner DiT blocks (linear layers, MLP, normalization) operate on the packed tensor at no extra cost. After the main network, we unpack back to a \((B, M_{\max}, D)\) tensor for de-chunking. This eliminates wasted FLOPs on padding tokens and keeps the realized inference cost proportional to the average compression ratio rather than the worst-case per-batch token count.

\subsection{De-chunking Layer}
\label{sec:dechunking}
After the inner network processes the shortened sequence, we reconstruct the token sequence back to its original resolution via a de-chunking layer with two conceptual components: \textbf{smoothing} over boundary-token representations and a \textbf{plug-back} map that assigns each original token position a boundary-derived representation.

\textbf{Motivation for smoothing.}
Hard keep/drop decisions can make chunk assignments unstable: small changes in router probabilities may shift a boundary and abruptly reassign many positions to a different retained token. This is especially common early in training, when probabilities often lie near the threshold. To improve stability, we smooth the reconstructed representation using the router's confidence rather than relying solely on the hard mask. High-confidence retained tokens are treated as reliable representatives, while low-confidence retained tokens are blended with nearby retained tokens, reducing discontinuities when the router is uncertain.

\textbf{Spatial smoothing.}
Let \(\mathcal{S}=\{s_1,\ldots,s_M\}\) be the retained boundary indices after chunking. For each retained token \(s_i\), let \(\mathbf{h}_i \in \mathbb{R}^D\) denote its representation after the DiT backbone, \(\mathbf{u}_i=(r_i,c_i)\) its original 2D grid coordinate, and \(p_i\) its router probability. Smoothing operates only over these \(M\) retained tokens. We compute pairwise squared distances \(d_{ij}^2=\|\mathbf{u}_i-\mathbf{u}_j\|_2^2\) and use a Gaussian kernel weighted by the source token's confidence:
\[
W_{ij} = \exp\!\left(-\frac{d_{ij}^2}{2\sigma^2}\right) \cdot p_j, \qquad
\tilde{W}_{ij} = \frac{W_{ij}}{\sum_{k} W_{ik}}.
\]
For each retained token, we compute a neighborhood-smoothed representation \(\tilde{\mathbf{h}}_i = \sum_j \tilde{W}_{ij} \mathbf{h}_j\) and blend it with the original representation according to the target token's confidence:
\begin{equation}
\mathbf{h}_i^{\mathrm{out}} = p_i \, \mathbf{h}_i + (1 - p_i) \, \tilde{\mathbf{h}}_i.
\end{equation}
High-confidence boundaries retain their original features, while low-confidence boundaries are smoothed toward their spatial neighbors. This confidence-weighted blend makes uncertain representatives less sensitive to a single hard routing decision: they borrow context from nearby retained tokens, whereas confident representatives are passed through with little change.

The \textbf{plug-back} map then reconstructs the full \(L\)-token grid by assigning each original grid position the representation of its spatially nearest boundary (Euclidean distance on the 2D grid). Thus, each dropped token receives the smoothed representation of its closest retained representative, while retained tokens plug back their own smoothed features at their original grid locations.

\subsection{Training objective and multi-budget training}
We train DC-DiT with the same diffusion training objective as in DiT \cite{peebles2023scalablediffusionmodelstransformers}. In addition, we include a lightweight regularizer following the load balancing mechanism of Mixture-of-Experts models \cite{fedus2022switch} that encourages a target average downsampling factor for the routing module. We denote this target compression ratio by \(N>1\); \(N\) is a training hyperparameter that specifies the desired average ratio between the original sequence length and the retained boundary-token length. Concretely, for a routing module output with boundary mask \(m \in \{0,1\}^{B \times L}\) and boundary probabilities \(p \in [0,1]^{B \times L}\), we define \(\hat{r} = \mathbb{E}[m]\) and \(\bar{p} = \mathbb{E}[p]\). We use the following regularizer:
\[
\mathcal{L}_{ratio} = \frac{N}{N-1}\left((1-\hat{r})(1-\bar{p}) + (N-1)\hat{r}\bar{p}\right).
\]

\textbf{Multi-budget training.}
Applying tail dropping only at inference would create a train--test mismatch: the inner DiT blocks would be trained only on the router's natural budget and then evaluated on more aggressively compressed sequences. We therefore train DC-DiT across several tail-dropping settings. Let \(\mathcal{R} = \{\rho_1, \rho_2, \ldots, \rho_K\}\) denote a fixed set of tail-dropping fractions, including \(\rho=0\). Conceptually, we optimize the average diffusion objective over these budgets,
\[
\mathcal{L}_{\mathrm{mb}} = \frac{1}{K}\sum_{\rho \in \mathcal{R}} \mathcal{L}^{(\rho)}_{\mathrm{diffusion}},
\]
where \(\mathcal{L}^{(\rho)}_{\mathrm{diffusion}}\) is the standard diffusion loss evaluated after applying tail dropping with fraction \(\rho\). In practice, after an initial warmup phase we optimize an unbiased stochastic estimate of this objective by sampling one \(\rho \sim \mathcal{R}\) per iteration and applying exactly the same tail-dropping path used at inference. The ratio loss \(\mathcal{L}_{ratio}\), however, is always computed on the router's natural boundary set before tail dropping, which keeps the router anchored to the target compression factor \(N\) while \(\rho\) provides an additional inference-time budget on top of the learned routing policy. The resulting training objective is
\[
\mathcal{L} = \mathcal{L}_{\mathrm{mb}} + \lambda \mathcal{L}_{ratio}.
\]

\begin{figure}[t]
\centering
\includegraphics[width=\textwidth]{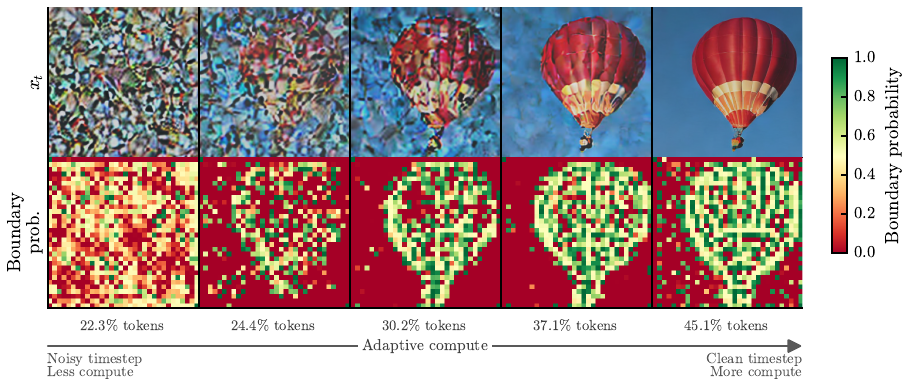}
\caption{Adaptive compute allocation learned by DC-DiT. Across diffusion timesteps, the router retains boundary tokens in spatially informative regions such as object structure, edges, and texture, while compressing more predictable background regions. The retained-token pattern changes over the denoising trajectory, allocating compute differently as images evolve from noisy structure to fine detail.}
\label{fig:adaptive_compute}
\end{figure}

\section{Experiments and Results}
\label{sec:experiments}

We evaluate DC-DiT on class-conditional ImageNet generation at 256px and 512px resolutions and compare quality--compute tradeoff against DiT baselines that use fixed patchification. Furthermore, we evaluate DC-DiT's composability with other dynamic computation and FLOPs-saving techniques such as DyDiT~\cite{wang2024dynamicdiffusiontransformer} and TeaCache~\cite{liu2025timestepembeddingteacache}. Additionally, we demonstrate lightweight upcycling of the text-to-image foundation model Z-Image~\cite{zimage2025efficient} into a DC-DiT to enable elastic inference while preserving high-quality image generation and prompt following.

\subsection{Experimental setup}
For our primary experiments on class-conditional ImageNet generation, we report FID-50K as our generation quality metrics. We use the same diffusion formulation as standard DiT~\cite{peebles2023scalablediffusionmodelstransformers}: a linear noise schedule with 1000 diffusion steps during training, and DDPM sampling with 250 steps. All models operate in the latent space of a pretrained Stable Diffusion VAE encoder~\cite{Rombach_2022_CVPR}, with class conditioning via adaLN-Zero.

\textbf{Model configurations.} We train DC-DiT at multiple model scales corresponding to the S, B, L, and XL variants of DiT. For each scale, we keep the transformer backbone identical to the corresponding DiT baseline and augment it with the encoder-router-decoder scaffold. This scaffold adds a small FLOP overhead relative to the matched DiT backbone, but unlike fixed-patch DiT, DC-DiT can be evaluated across a range of inference budgets by varying the tail-dropping fraction \(\rho\). We train DC-DiT with target compression ratio \(N{=}4\), corresponding to patch size \(P{=}2\) in standard DiT.

\textbf{Training.} All models are trained with a global batch size of 256 using AdamW with learning rate \(1{\times}10^{-4}\). The DiT baselines are trained for 400K steps. For DC-DiT, we use multi-budget training and sample the tail-dropping fraction \(\rho\) from \(\{0.0, 0.1, 0.2, 0.3, 0.4, 0.5, 0.6\}\). Because tail dropping reduces the effective FLOPs of a DC-DiT training step, we extend DC-DiT training so that its total training compute matches the corresponding baseline budget (details in Appendix~\ref{app:multi-budget-training}). We set the ratio-loss weight to \(\lambda{=}0.03\) based on a grid search. All models are trained on AMD Instinct MI325X and MI300X GPUs.

\subsection{Main Results}

Table~\ref{tab:main_results} presents the main ImageNet results across model scales, guidance settings, and resolutions. Across these settings, DC-DiT improves the FID--FLOPs Pareto frontier relative to fixed-patch DiT baselines: near the DiT compute budget, dynamic chunking yields better FID, while more aggressive tail dropping exposes substantially cheaper operating points from the same checkpoint. The benefits of DC-DiT are emphasized under Lite-CFG, where DC-DiT improves FID by up to \(37.8\%\) while reducing inference compute, and at \(512{\times}512\), where DC-DiT-XL reduces compute by \(36.8\%\) with only a small FID tradeoff. Overall, we observe highly meaningful gains in compute reduction and FID scores for the S, B and L model scales while the XL model scale achieves competitive performance to baseline DiT while utilizing lower FLOPs.

\begin{table}[t]
\centering
\caption{Main ImageNet results across model scale, guidance setting, and resolution. In Lite-CFG rows, the tail-drop value applies to the unconditional branch.}
\label{tab:main_results}
\scriptsize
\setlength{\tabcolsep}{1.6pt}
\begin{minipage}[t]{0.49\textwidth}
\vspace{0pt}
\centering
\begin{tabular}{@{}llcccc@{}}
\toprule
Scale & Model & Params (M) & Tail drop & TFLOPs/img \(\downarrow\) & FID-50K \(\downarrow\) \\
\midrule
\rowcolor{mygold!30}
\multicolumn{6}{c}{\textbf{\textit{\(256{\times}256,\) no classifier-free guidance}}} \\
S & DiT-S/2 & 33 & -- & 3.02 & 68.40 \\
\rowcolor{mycolor2!70}
S & DC-DiT-S & 34.7 & 0.2 & 3.06 & \textbf{58.43} \textcolor{green!50!black}{(-14.6\%)} \\
\rowcolor{mycolor2!70}
S & DC-DiT-S & 34.7 & 0.5 & \textbf{2.14} \textcolor{green!50!black}{(-29.1\%)} & 66.56 \\
\midrule
B & DiT-B/2 & 131 & -- & 11.46 & 43.47 \\
\rowcolor{mycolor2!70}
B & DC-DiT-B & 137 & 0.2 & 11.94 & \textbf{33.91} \textcolor{green!50!black}{(-22.0\%)} \\
\rowcolor{mycolor2!70}
B & DC-DiT-B & 137 & 0.5 & \textbf{8.55} \textcolor{green!50!black}{(-25.4\%)} & 44.64 \\
\midrule
L & DiT-L/2 & 459 & -- & 40.21 & 23.33 \\
\rowcolor{mycolor2!70}
L & DC-DiT-L & 469 & 0.1 & 39.61 & \textbf{22.32} \textcolor{green!50!black}{(-4.3\%)} \\
\rowcolor{mycolor2!70}
L & DC-DiT-L & 469 & 0.3 & \textbf{31.86} \textcolor{green!50!black}{(-20.8\%)} & 23.94 \\
\midrule
XL & DiT-XL/2 & 675 & -- & 59.00 & \textbf{19.47} \\
\rowcolor{mycolor2!70}
XL & DC-DiT-XL & 689 & 0 & 62.99 & 19.73 \\
\rowcolor{mycolor2!70}
XL & DC-DiT-XL & 689 & 0.2 & 51.60 & 20.45 \\
\rowcolor{mycolor2!70}
XL & DC-DiT-XL & 689 & 0.4 & \textbf{40.26} \textcolor{green!50!black}{(-31.8\%)} & 22.52 \\
\midrule
\rowcolor{mygold!30}
\multicolumn{6}{c}{\textbf{\textit{\(512{\times}512,\) Lite-CFG with CFG \(=1.25\)}}} \\
XL & DiT-XL/2 & 675 & -- & 522.50 & \textbf{11.04} \\
\rowcolor{mycolor2!70}
XL & DC-DiT-XL & 693 & 0.9 & \textbf{330.00} \textcolor{green!50!black}{(-36.8\%)} & 12.50 \\
\bottomrule
\end{tabular}
\end{minipage}
\hfill
\begin{minipage}[t]{0.49\textwidth}
\vspace{0pt}
\centering
\begin{tabular}{@{}llcccc@{}}
\toprule
Scale & Model & Params (M) & Tail drop & TFLOPs/img \(\downarrow\) & FID-50K \(\downarrow\) \\
\midrule
\rowcolor{mygold!30}
\multicolumn{6}{c}{\textbf{\textit{\(256{\times}256,\) Lite-CFG with CFG \(=1.25\)}}} \\
S & DiT-S/2 & 33 & -- & 6.04 & 54.98 \\
\rowcolor{mycolor2!70}
S & DC-DiT-S & 34.7 & 0.9 & \textbf{4.73} \textcolor{green!50!black}{(-21.7\%)} & \textbf{38.99} \textcolor{green!50!black}{(-29.1\%)} \\
\midrule
B & DiT-B/2 & 131 & -- & 22.93 & 29.72 \\
\rowcolor{mycolor2!70}
B & DC-DiT-B & 137 & 0.9 & \textbf{18.38} \textcolor{green!50!black}{(-19.8\%)} & \textbf{18.49} \textcolor{green!50!black}{(-37.8\%)} \\
\midrule
L & DiT-L/2 & 459 & -- & 80.41 & 12.59 \\
\rowcolor{mycolor2!70}
L & DC-DiT-L & 469 & 0.9 & \textbf{52.88} \textcolor{green!50!black}{(-34.2\%)} & \textbf{10.17} \textcolor{green!50!black}{(-19.2\%)} \\
\midrule
XL & DiT-XL/2 & 675 & -- & 118.13 & 9.45 \\
\rowcolor{mycolor2!70}
XL & DC-DiT-XL & 689 & 0.9 & \textbf{76.13} \textcolor{green!50!black}{(-35.6\%)} & \textbf{8.80} \textcolor{green!50!black}{(-6.9\%)} \\
\midrule
\rowcolor{mygold!30}
\multicolumn{6}{c}{\textbf{\textit{\(512{\times}512,\) no classifier-free guidance}}} \\
B & DiT-B/2\textsuperscript{\(\dagger\)} & 131 & -- & 53.09 & 49.58 \\
\rowcolor{mycolor2!70}
B & DC-DiT-B & 139 & 0.2 & 52.60 & \textbf{42.64} \textcolor{green!50!black}{(-14.0\%)} \\
\rowcolor{mycolor2!70}
B & DC-DiT-B & 139 & 0.5 & \textbf{35.94} \textcolor{green!50!black}{(-32.3\%)} & 48.40 \\
\midrule
XL & DiT-XL/2 & 675 & -- & 261.25 & \textbf{20.13} \\
\rowcolor{mycolor2!70}
XL & DC-DiT-XL & 693 & 0.2 & \textbf{228.75} \textcolor{green!50!black}{(-12.4\%)} & 23.59 \\
\bottomrule
\end{tabular}
\end{minipage}
\end{table}
\begingroup
\renewcommand{\thefootnote}{\fnsymbol{footnote}}
\footnotetext[2]{At 512px, DC-DiT-S and DC-DiT-L trained stably, whereas the corresponding vanilla DiT-S/2 and DiT-L/2 baselines repeatedly encountered loss divergence despite multiple restarts.}
\endgroup

\textbf{Learned spatio-temporal compression.}
Figure~\ref{fig:adaptive_compute} visualizes the router's boundary predictions on a representative ImageNet sample. Spatially, the router assigns high boundary probability to object edges, fine textures, and regions of high local variation, while dropping tokens in uniform backgrounds and other predictable areas. Temporally, it retains fewer tokens at early, noisier denoising steps and more tokens at later steps, when fine details are resolved. Thus, DC-DiT learns an implicit content- and timestep-adaptive tokenization: simple background regions and noisy early states are compressed more aggressively, whereas detailed object regions and later denoising stages receive more tokens. This behavior emerges solely from the diffusion training objective, without explicit supervision for segmentation, boundary detection, or timestep-dependent compute scheduling. It is consistent with the coarse-to-fine allocation studied in prior work on elastic visual generation, such as ELIT~\cite{haji2026one} and MaGNeTS~\cite{goyal2025masked}, but arises naturally rather than being manually prescribed.

\subsection{Ablations}

\begin{wrapfigure}{r}{0.52\linewidth}
    \centering
    \vspace{-0.5em}
    \includegraphics[width=\linewidth]{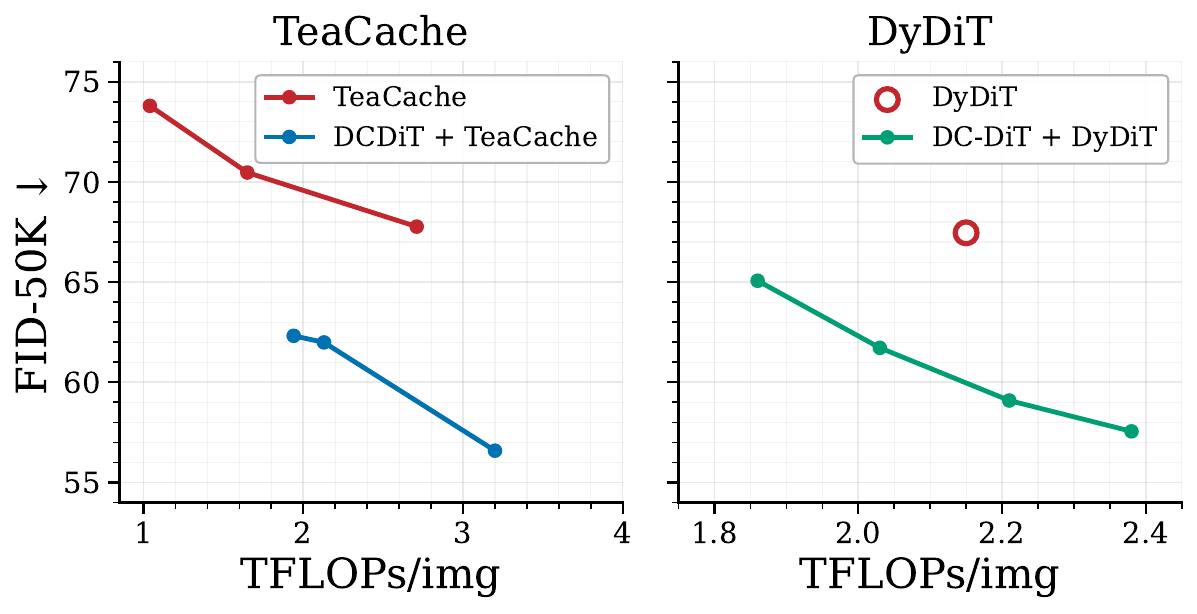}
    \vspace{-1.0em}
    \caption{DCDiT is complementary to TeaCache and DyDiT, improving FID at comparable compute budgets.}
    \label{fig:dcdit_orthogonal}
    \vspace{-0.8em}
\end{wrapfigure}
We ablate DC-DiT's main components and evaluate robustness under increasingly aggressive tail dropping. Table~\ref{tab:ablation_components} show that multi-budget training, the spatial predictability router, and de-chunk smoothing all improve generation quality. The spatial predictability router outperforms a direct 2D adaptation of the H-Net cosine-similarity router~\cite{hnet}. Finally, removing de-chunk smoothing worsens FID, consistent with its role in stabilizing hard routing decisions and improving the quality of the learned segmentations. Multi-budget training is particularly important for elastic inference: by randomly dropping low-confidence retained tokens during training, it encourages reconstruction-critical tokens to appear earlier in the router ranking, leading to more graceful degradation as the inference drop ratio $\rho$ increases and slightly better FID even at $\rho=0$ as shown in Figure~\ref{fig:tail_dropping_ablation}. 

\subsection{Upcycling Z-Image to DC-Z-Image}

\begin{figure}[t]
\centering
\begin{minipage}[t]{0.57\textwidth}
\vspace{0pt}
\centering
\includegraphics[width=\linewidth]{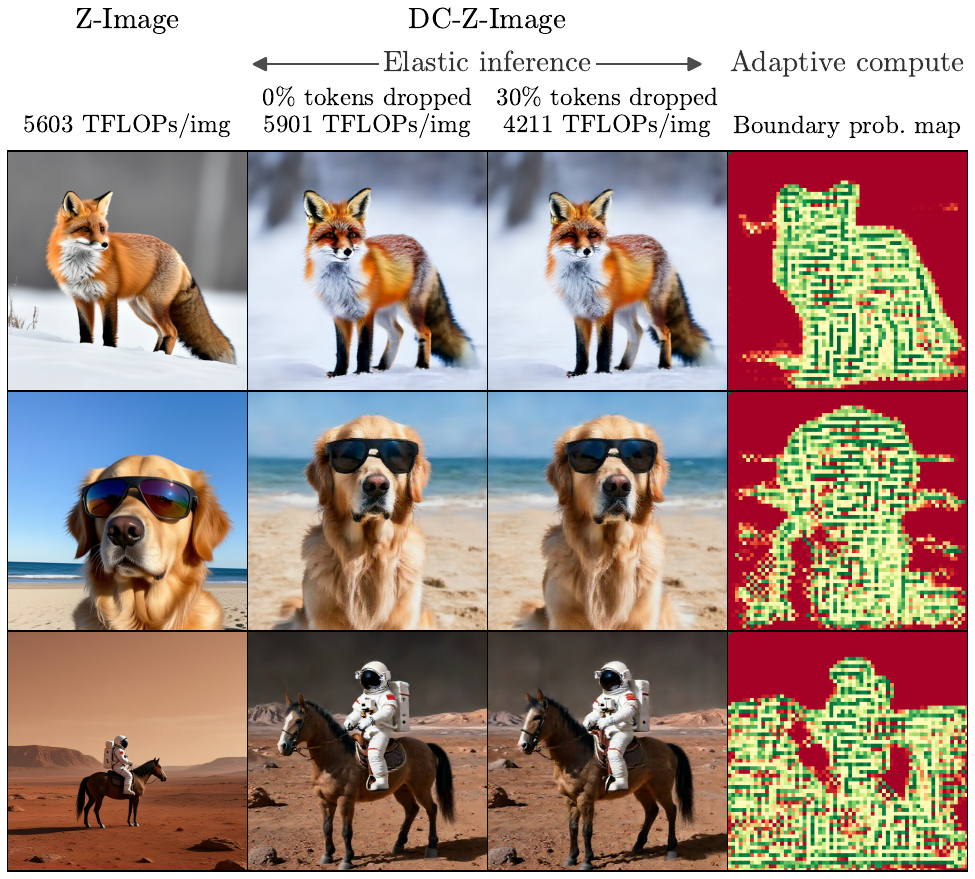}
\caption{Qualitative elastic inference examples after upcycling Z-Image. DC-Z-Image preserves high-quality, prompt-aligned generations across compute budgets.}
\label{fig:zimage_qualitative}
\end{minipage}
\hfill
\begin{minipage}[t]{0.39\textwidth}
\vspace{0pt}
\centering
\setlength{\tabcolsep}{2pt}
\begin{tabular}{@{}lcc@{}}
\toprule
Ablation & TFLOPs/img & FID-50K \\
\midrule
DC-DiT-S & 3.69 & \textbf{56.73} \\
No multi-budget training & 3.67 & 57.77 \\
Cosine similarity router & 3.67 & 57.96 \\
No dechunk smoothing & 3.47 & 59.79 \\
\bottomrule
\end{tabular}
\refstepcounter{table}\label{tab:ablation_components}
\vspace{2pt}

\noindent\parbox{\linewidth}{\small\textbf{Table~\thetable:} Component ablations on DC-DiT-S. Removing any critical component meaningfully worsens FID.}

\vspace{0.8em}
\includegraphics[width=\linewidth]{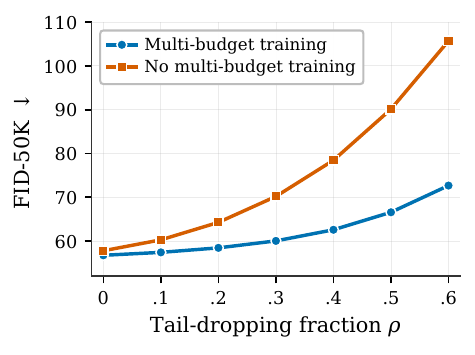}
\vspace{-2em}
\caption{Tail-dropping robustness on DC-DiT-S. Multi-budget training keeps FID stable as \(\rho\) increases.}
\label{fig:tail_dropping_ablation}
\end{minipage}
\end{figure}

To test whether dynamic chunking scales beyond class-conditional ImageNet models, we apply DC to Z-Image~\cite{zimage2025efficient}, a state-of-the-art text-to-image diffusion-transformer trained with flow matching. Instead of training from scratch, we upcycle the Z-Image model to DC-Z-Image by replacing fixed patchification with the encoder-router-decoder scaffold and performing lightweight adaptation. Our upcycling uses 5K steps of distillation following the grafting-style adaptation outlined in~\cite{chandrasegaran2025grafting} and 10K flow-matching steps on 1M synthetic Z-Image samples with prompts from Recap-DataComp-1B~\cite{li2024recaption}. During upcycling, we freeze the timestep and text-embedding modules and add a trainable LayerNorm adaptor to the encoder/decoder conditioning vector. As shown in Table~\ref{tab:zimage_dpg} and Figure~\ref{fig:zimage_qualitative}, DC-Z-Image preserves prompt-following quality across elastic inference budgets: increasing tail dropping reduces cost from $5901$ to $2570$ TFLOPs/img, while maintaining DPG-Bench scores comparable to the original Z-Image baseline.

\begin{table}[t]
\centering
\caption{DPG-Bench preservation after upcycling Z-Image. DC-Z-Image preserves prompt-following quality across elastic inference budgets while reducing inference compute at higher tail-dropping fractions.}
\label{tab:zimage_dpg}
\scriptsize
\resizebox{\textwidth}{!}{%
\begin{tabular}{@{}lcccccccc@{}}
\toprule
Model & Tail drop & TFLOPs/img & Relation & Entity & Other & Attribute & Global & DPG-Bench \\
\midrule
Z-Image & -- & 5603 & 92.72 & 89.95 & 90.02 & 91.08 & \textbf{93.06} & 85.97 \\
DC-Z-Image & 0 & 5901 & \textbf{93.15} & \textbf{92.38} & 89.47 & 90.72 & 91.03 & 86.82 \\
DC-Z-Image & 30\% & 4211 & 92.89 & 91.70 & 88.15 & \textbf{92.02} & 89.34 & \textbf{87.64} \\
DC-Z-Image & 60\% & 2570 & 92.03 & 90.79 & \textbf{92.91} & 89.23 & 91.64 & 86.32 \\
\bottomrule
\end{tabular}%
}
\end{table}

\subsection{Composability with other dynamic computation techniques }
DC-DiT introduces content-adaptive patchification while leaving the DiT backbone unchanged, making it compatible with orthogonal dynamic execution methods~\cite{wang2024dynamicdiffusiontransformer,wu2025importancebasedtokenmergingefficient,liu2025timestepembeddingteacache}. To demonstrate composability, we combine DC-DiT with DyDiT~\cite{wang2024dynamicdiffusiontransformer} and TeaCache~\cite{liu2025timestepembeddingteacache}. DyDiT adds lightweight learned gates to modulate backbone computation across timesteps, while TeaCache is a training-free method that reuses model outputs across denoising steps. We apply both methods to the DiT backbone inside DC-DiT to further reduce FLOPs. As shown in Figure~\ref{fig:dcdit_orthogonal}, DC-DiT remains compatible with both approaches, achieving additional compute reductions while preserving or improving FID.

\section{Conclusion}
We introduced DC-DiT, a diffusion transformer that replaces fixed patchification with adaptive tokenization learned end-to-end through diffusion training. Its router reallocates compute across spatial regions and denoising timesteps without explicit supervision, while multi-budget training turns the same routing signal into elastic inference and Lite-CFG. Across ImageNet settings, DC-DiT improves the quality--compute Pareto frontier, reducing inference FLOPs by up to $36.8\%$ and improving FID by up to $37.8\%$. The approach also upcycles Z-Image with lightweight adaptation and composes well with DyDiT and TeaCache. These results position adaptive tokenization as a practical primitive for efficient diffusion models.

\bibliographystyle{unsrtnat}
\bibliography{references}  

\appendix

\section{Additional Visual Results and Reproducibility Details}

\subsection{Additional visual results}
\begin{figure}[!htbp]
\centering
\includegraphics[width=\textwidth,height=0.72\textheight,keepaspectratio]{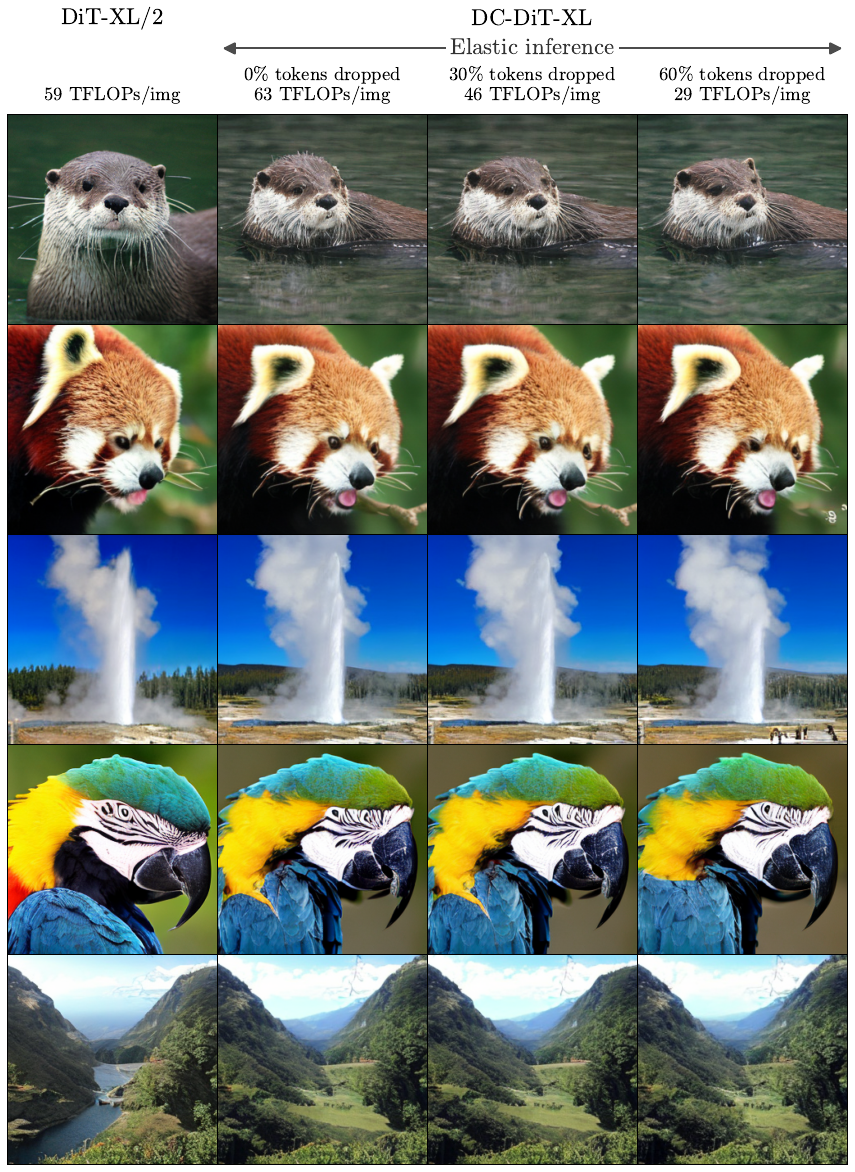}
\caption{Extended qualitative elastic inference results. This figure expands the tail-dropping grid from Figure~\ref{fig:main_results}, showing additional generations from the same checkpoint across user-selected compute budgets.}
\label{fig:tail_dropping_grid_appendix}
\end{figure}

\begin{figure}[!htbp]
\centering
\includegraphics[width=\textwidth,height=0.72\textheight,keepaspectratio]{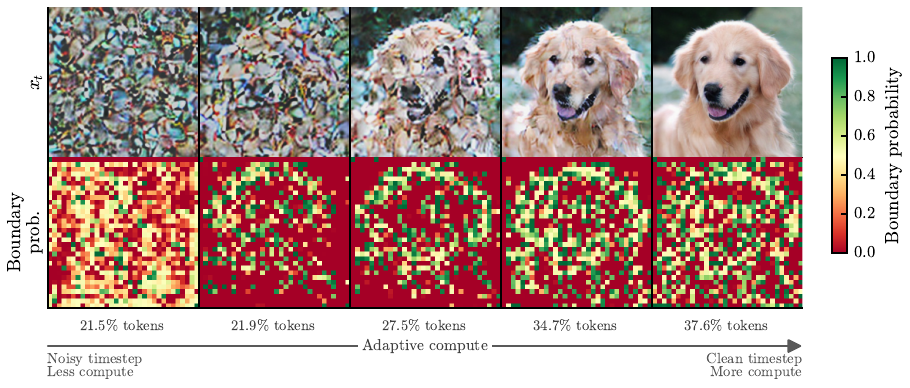}
\caption{Extended adaptive compute visualization. This figure expands Figure~\ref{fig:adaptive_compute} with additional examples of the router's learned boundary predictions across diffusion timesteps.}
\label{fig:chunking_viz_appendix}
\end{figure}

Figure~\ref{fig:tail_dropping_grid_appendix} provides an extended version of the qualitative elastic inference grid shown in Figure~\ref{fig:main_results}. Figure~\ref{fig:chunking_viz_appendix} provides an extended version of the adaptive compute visualization shown in Figure~\ref{fig:adaptive_compute}, with additional samples illustrating the learned spatial chunking and timestep-dependent token allocation.

\subsection{FLOPs accounting}
\label{app:flops-accounting}
We report TFLOPs/img as the total floating-point operation count required to generate one image with the sampling protocol used for evaluation. For unguided ImageNet sampling, this is the sum of the per-forward cost across the 250 DDPM sampling steps. For standard classifier-free guidance, each denoising step includes both conditional and unconditional model evaluations. For Lite-CFG, we count these two branches separately, using the conditional branch cost at its conservative tail-dropping budget and the unconditional branch cost at the more aggressive tail-dropping budget reported in Table~\ref{tab:main_results}.

For each model forward pass, we count all matrix multiplications, convolutions, normalizations, elementwise operations, embeddings, and output projections. This includes the overhead introduced by the encoder-router-decoder scaffold. Consequently, the reported FLOPs reflect the realized end-to-end cost of DC-DiT at a given tail-dropping fraction.

Variable-length sequence packing is accounted for using the actual retained sequence length of each sample. Let \(L_b\) denote the number of retained tokens for sample \(b\), \(H\) the number of attention heads, and \(d_h\) the head dimension. Pointwise components of the packed DiT blocks, such as linear layers, MLPs, and normalization, scale with \(\sum_b L_b\). Self-attention is counted using the packed variable-length cost
\[
H \sum_b L_b^2 (4d_h + 3),
\]
rather than the padded cost \(BH M_{\max}^2(4d_h+3)\), where \(M_{\max}=\max_b L_b\). This distinction is important because DC-DiT samples in a batch may retain different numbers of tokens; padding to the longest sequence would overestimate the cost and would not match the packed attention computation used during inference. For fixed-patch DiT baselines, \(L_b\) is constant across the batch and the expression reduces to the standard dense attention cost.

\subsection{Multi-budget training and compute matching}
\label{app:multi-budget-training}
DC-DiT uses multi-budget training so that a single checkpoint can be evaluated at several tail-dropping fractions. We first train for a 5K-step warmup with no tail dropping, which lets the router converge toward the target compression ratio before exposing the inner DiT blocks to more aggressively compressed sequences. After this warmup, each training step samples \(\rho\) uniformly from \(\mathcal{R}=\{0.0,0.1,0.2,0.3,0.4,0.5,0.6\}\) and applies the same tail-dropping path used at inference.

Because tail dropping changes the cost of a DC-DiT training step, we choose the total number of DC-DiT steps to match the training compute of the corresponding 400K-step DiT baseline. Let \(S_{\mathrm{DiT}}=400{,}000\), \(S_{\mathrm{warm}}=5{,}000\), \(F_{\mathrm{DiT}}\) be the measured DiT FLOPs per image, \(F_0\) be the measured DC-DiT FLOPs per image during warmup, and \(\bar{F}_{\mathrm{mb}}\) be the average measured DC-DiT FLOPs per image over the sampled tail-dropping budgets in \(\mathcal{R}\). We set
\[
S_{\mathrm{DC}} =
S_{\mathrm{warm}} +
\frac{S_{\mathrm{DiT}}F_{\mathrm{DiT}} - S_{\mathrm{warm}}F_0}
{\bar{F}_{\mathrm{mb}}},
\]
rounded to the nearest integer. Table~\ref{tab:compute_matched_steps} reports the resulting training lengths for the ImageNet settings that have corresponding matched DiT baselines in Table~\ref{tab:main_results}.

\begin{table}[!htbp]
\centering
\caption{Compute-matched training lengths for the ImageNet DC-DiT experiments. DC-DiT trains longer than 400K steps when multi-budget tail dropping reduces the average cost per post-warmup step.}
\label{tab:compute_matched_steps}
\small
\begin{tabular}{@{}llccc@{}}
\toprule
Resolution & Scale & DiT TFLOPs/img & DiT steps & DC-DiT steps \\
\midrule
\(256{\times}256\) & S  & 3.02   & 400K & 421{,}230 \\
\(256{\times}256\) & B  & 11.46  & 400K & 408{,}071 \\
\(256{\times}256\) & L  & 40.21  & 400K & 486{,}372 \\
\(256{\times}256\) & XL & 59.00  & 400K & 494{,}943 \\
\(512{\times}512\) & B  & 53.09  & 400K & 432{,}097 \\
\(512{\times}512\) & XL & 261.25 & 400K & 495{,}365 \\
\bottomrule
\end{tabular}
\end{table}

\subsection{Architecture details by scale}
Table~\ref{tab:architecture_by_scale} summarizes the basic shape parameters for the DC-DiT scale configurations used in the paper. The encoder/decoder bottleneck width is computed from the configured hidden width and dimension reduction factor.

\begin{table}[!htbp]
\centering
\caption{Basic architecture shapes for the DC-DiT scale configurations.}
\label{tab:architecture_by_scale}
\small
\begin{tabular}{@{}lcccccc@{}}
\toprule
Configuration & Enc. blocks & Transformer blocks & Dec. blocks & Hidden width & Heads & Enc./Dec. bottleneck \\
\midrule
DC-DiT-S  & 2 & 12 & 2 & 384  & 6  & 96 \\
DC-DiT-B  & 2 & 12 & 2 & 768  & 12 & 192 \\
DC-DiT-L  & 2 & 24 & 2 & 1024 & 16 & 256 \\
DC-DiT-XL & 2 & 28 & 2 & 1152 & 16 & 288 \\
\bottomrule
\end{tabular}
\end{table}

\subsection{Hyperparameters}
Table~\ref{tab:repro_hyperparams} lists the main hyperparameters used for the ImageNet DC-DiT-B \(N=4\) experiments. Unless otherwise noted, the same optimizer, diffusion, routing, and multi-budget settings are used across ImageNet model scales.

\begin{table}[!htbp]
\centering
\caption{Main hyperparameters for ImageNet DC-DiT training.}
\label{tab:repro_hyperparams}
\small
\begin{tabular}{@{}ll@{}}
\toprule
Hyperparameter & Value \\
\midrule
Dataset & ImageNet, 1000 classes \\
Latent resolution & \(256{\times}256\) \\
Diffusion schedule & Linear, 1000 training steps \\
Sampler & DDPM, 250 sampling steps \\
Global batch size & 256 \\
Optimizer & AdamW \\
Learning rate & \(1{\times}10^{-4}\) \\
Weight decay & 0 \\
EMA decay & 0.9999 \\
Gradient clipping & 1.0 \\
Classifier-free dropout & 0.1 \\
Variance prediction & Learned \(\sigma\) \\
Target compression & \(N=4\) \\
Ratio loss weight & 0.03 \\
Ratio-loss batch size & 16 \\
Router & Spatial predictability \\
De-chunk smoothing & Enabled, Gaussian \(\sigma=1.0\) \\
Multi-budget warmup & 5K steps, no tail dropping \\
Tail-drop fractions & \(\{0.0,0.1,0.2,0.3,0.4,0.5,0.6\}\) \\
\bottomrule
\end{tabular}
\end{table}

\end{document}